\documentclass[letterpaper, 10pt, conference]{ieeeconf/ieeeconf}    

\makeatletter
\let\NAT@parse\undefined
\makeatother

\usepackage[numbers,sectionbib,sort]{natbib}

\let\labelindent\relax
\usepackage{enumitem}

\usepackage{bm}
\usepackage{gensymb}
\usepackage{graphicx}
\usepackage{amsmath}
\usepackage{algorithm}
\usepackage[font=small,skip=0pt]{subcaption}
\usepackage{amsfonts}
\usepackage{siunitx}
\usepackage{booktabs}
\usepackage[font=small]{caption}
\usepackage[export]{adjustbox}
\usepackage{sidecap} \sidecaptionvpos{figure}{c}
\usepackage{multirow}
\usepackage{algorithmic}
\usepackage{comment}

\usepackage{hyperref}
\hypersetup{colorlinks,breaklinks,
linkcolor=[rgb]{0.5,0.,0.},
citecolor=[rgb]{0.000,0.427,0.173},
urlcolor=[rgb]{0.031,0.318,0.612}}

\usepackage[printonlyused,withpage,nolist,nohyperlinks]{acronym}
\begin{acronym}

\acro{2D}{two-dimensional}

\acro{3D}{three-dimensional}

\acro{AHRS}{attitude and heading reference system}

\acro{AUV}{autonomous underwater vehicle}

\acro{CPP}{Chinese Postman Problem}

\acro{DoF}{degree of freedom}
\acrodefplural{DoF}[DoFs]{degrees of freedom}

\acro{DVL}{Doppler velocity log}

\acro{FSM}{finite state machine}

\acro{IMU}{inertial measurement unit}

\acro{LBL}{Long Baseline}

\acro{MCM}{mine countermeasures}

\acro{MDP}{Markov decision process}
\acrodefplural{MDP}[MDPs]{Markov decision processes}

\acro{POMDP}{partially observable Markov decision process}
\acrodefplural{POMDP}[POMDPs]{partially observable Markov decision processes}

\acro{PRM}{Probabilistic Roadmap}
\acrodefplural{PRM}[PRM]{Probabilistic Roadmaps}

\acro{ROI}{region of interest}
\acrodefplural{ROI}[ROIs]{regions of interest}

\acro{ROS}{Robot Operating System}

\acro{ROV}{remotely operated vehicle}

\acro{RRT}{Rapidly-exploring Random Tree}
\acrodefplural{RRT}[RRTs]{Rapidly-exploring Random Trees}

\acro{SLAM}{simultaneous localization and mapping}

\acro{SSE}{sum of squared errors}

\acro{STOMP}{Stochastic Trajectory Optimization for Motion Planning}

\acro{TRN}{Terrain-Relative Navigation}

\acro{UAV}{Unmanned Aerial Vehicle}

\acro{UGV}{Unmanned Ground Vehicle}

\acro{MAV}{Micro Aerial Vehicle}

\acro{USBL}{Ultra-Short Baseline}

\acro{IPP}{informative path planning}

\acro{FoV}{field of view}
\acrodefplural{FoV}[FoVs]{fields of view}

\acro{CDF}{cumulative distribution function}

\acro{ML}{maximum likelihood}

\acro{RMSE}{Root Mean Squared Error}
\acro{MLL}{Mean Log Loss}

\acro{GP}{Gaussian Process}
\acrodefplural{GP}[GPs]{Gaussian processes}

\acro{KF}{Kalman Filter}
\acro{EKF}{Extended Kalman Filter}

\acro{IP}{Interior Point}
\acro{BO}{Bayesian Optimization}

\acro{GPS}{Global Positioning System}

\acro{PDM}{probability density map}
\acro{VIO}{Visual Inertial Odometry}
\acro{MSF}{Multi Sensor Fusion}
\acro{MPC}{model predictive control}
\acro{ToF}{time-of-flight}

\end{acronym}

\IEEEoverridecommandlockouts                           
\title{\LARGE \bf
UAV/UGV Autonomous Cooperation:\\
UAV assists UGV to climb a cliff by attaching a tether
}

\author{
    Takahiro Miki, Petr Khrapchenkov and Koichi Hori
\thanks{Authors are with the University of Tokyo, Japan.}\thanks{\tt\small takahiro.miki1992@gmail.com} \thanks{\tt\small petr.khrapchenkov@gmail.com} \thanks{\tt\small hori@computer.org}}

\begin{document}

\maketitle
\thispagestyle{empty}
\pagestyle{empty}

\begin{abstract}
This paper proposes a novel cooperative system for an \ac{UAV} and an \ac{UGV} which utilizes the \ac{UAV} not only as a flying sensor but also as a tether attachment device. 
Two robots are connected with a tether, allowing the \ac{UAV} to anchor the tether to a structure located at the top of a steep terrain, impossible to reach for \ac{UGV}s.
Thus, enhancing the poor traversability of the \ac{UGV} by not only providing a wider range of scanning and mapping from the air, but also by allowing the \ac{UGV} to climb steep terrains with the winding of the tether.
In addition, we present an autonomous framework for the collaborative navigation and tether attachment in an unknown environment. The \ac{UAV} employs visual inertial navigation with 3D voxel mapping and obstacle avoidance planning. The \ac{UGV} makes use of the voxel map and generates an elevation map to execute path planning based on a traversability analysis. 
Furthermore, we compared the pros and cons of possible methods for the tether anchoring from multiple points of view.
To increase the probability of successful anchoring, we evaluated the anchoring strategy with an experiment. 
Finally, the feasibility and capability of our proposed system were demonstrated by an autonomous mission experiment in the field with an obstacle and a cliff.
\end{abstract}

\section{INTRODUCTION}
\label{sec:introduction}

Recent advancements in the field of artificial intelligence and small, unmanned robots have lead to the growing range of applications. In the field of exploration robots, the ability to move through challenging environments, such as disaster areas, outdoor fields or planet surfaces is required.

The \ac{UAV}s have recently gained much interest among researchers and industries owing to its high accessibility.
They can go beyond obstacles, rough terrains or steep slopes, and are able to provide a view from a high altitude.

However, its payload or battery life is limited, and as a result, it becomes difficult to conduct missions that require heavy equipment or complex manipulations for the \ac{UAV}s.
On the other hand, the \ac{UGV}s have a higher battery capacity and larger payload, meaning it can carry heavier sensors, powerful computers and complex manipulators to perform actions.
In contrast, they often suffer from the limited sensor range, poor terrain traversability and climbing ability. 

To address the limited sensor range of the \ac{UGV}s, many researches for Air-Ground cooperation have been done.
\citet{michael2014collaborative} built a 3D map of an earthquake-damaged building with a team of a \ac{UAV} and a \ac{UGV}.
The \ac{UGV} carries the \ac{UAV} during the mission until they reach to a non-traversable area. When they arrive, the \ac{UAV} takes off and gathers the data to create a map.
Similarly, in \cite{hsieh2007adaptive, garzon2013aerial, guerin2015uav, fankhauser2016collaborative, shen2017collaborative, hood2017bird}, \ac{UAV}s and \ac{UGV}s perform a cooperative navigation or an exploration where \ac{UAV}s give a wider range of scanning for the \ac{UGV}s to navigate through the unknown environment.
To overcome the short battery life problem of \ac{UAV}s, tether powered drones were developed~\cite{choi2014tethered, zikou2015power, kiribayashi2018design}.
They provide power from the ground station via the tether to enable the \ac{UAV} to fly almost limitless.
Moreover, \citep{papachristos2014power, kiribayashi2015modeling} developed a tether connected \ac{UAV}/\ac{UGV} cooperation system. They can utilize the advantages of both flying and ground robots.

However, all of these cooperative approaches use the \ac{UAV}s only as a flying sensor that can provide a wider range of scanning.

\begin{figure}[t]
\begin{center}
   \includegraphics[width=7.0cm]{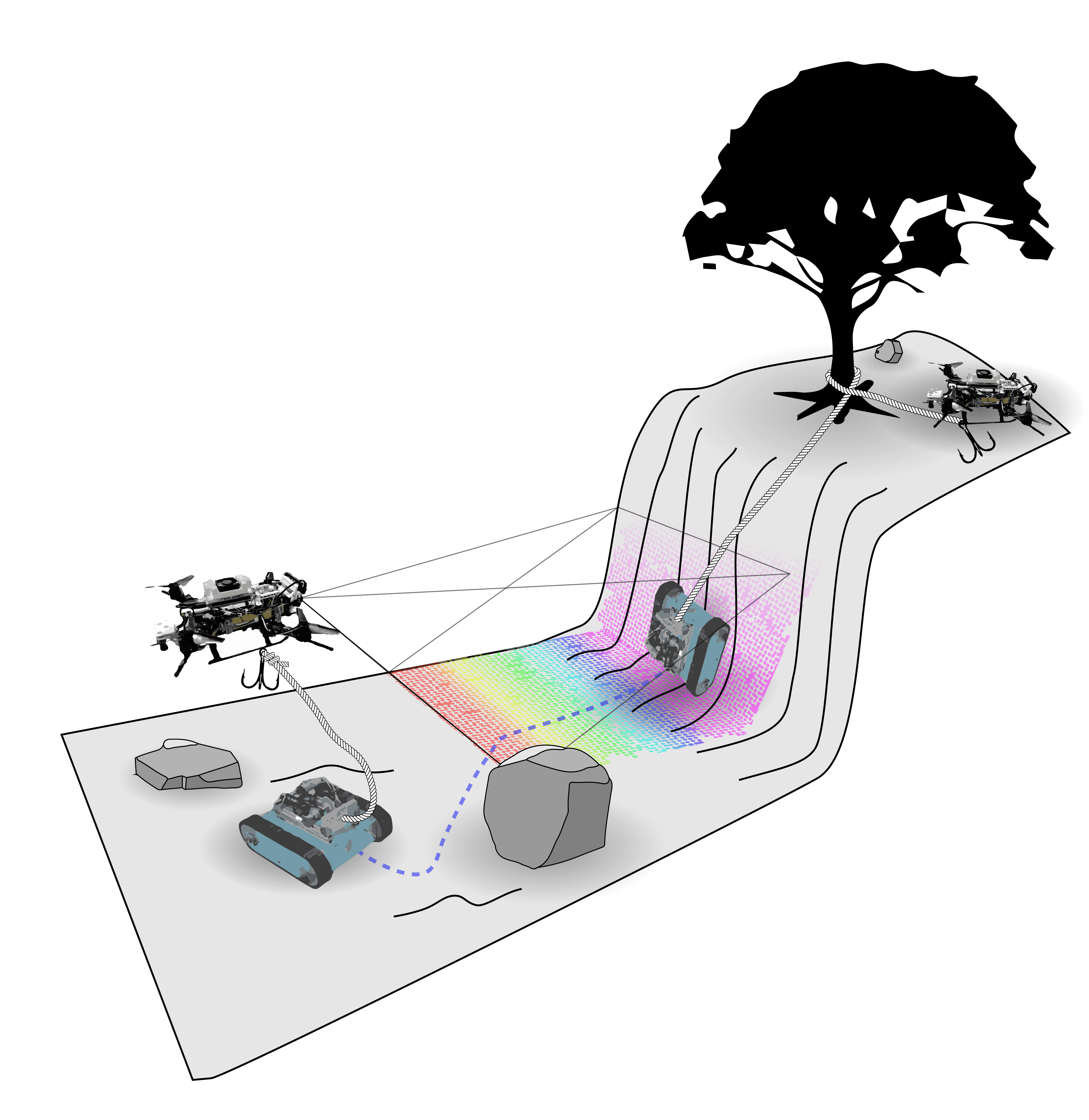}
  \end{center}
  \caption{Our proposed \ac{UAV}/\ac{UGV} cooperative system. The \ac{UAV} provides a sensor data for mapping. In addition, the \ac{UAV} attaches a tether to the structure on top of a cliff and the \ac{UGV} climbs by winding it.}
\end{figure}

To enhance the \ac{UGV}s traversability, one of the solutions is to use a tethered vehicle equipped with a winch.
Some off-road cars with a winch can climb a very steep slope by winding the tether anchored at the top of the slope.
Also, \citep{nesnasaxel, nesnas2008axel} developed a two-wheel rover which has a winch in the middle of the shaft. Their objective is to explore an inaccessible crater with a help of a mother vehicle.
\citep{walker2015qualification, walker2016update} also demonstrated the dual rover system connected with a tether to explore the possible skylights on the moon. A four-wheeled parent rover waits near the edge and the two-wheeled child rover explores the steep terrain.
\citep{mcgarey2016system, mcgarey2018field} made a tethered vehicle that can explore a steep terrain and create a 3D map of the environment.

Although they can enhance the rough and steep terrain traversability of the \ac{UGV}s, the tether is required to be anchored beforehand at the top of the slope by a parent rover or a human. In this case, the reachability will be limited to the already accessible area by the \ac{UGV} and area below, made possible by the tether.
As an exception, in \cite{drenner2002mobility}, a two-wheel rover equipped with a grappling hook, launches the hook to an object and use it as an anchor to raise itself. Nevertheless, it cannot climb over a cliff nor detect or observe the anchoring point unobservable from the lower ground.

In this research, we propose a \ac{UAV} / \ac{UGV} cooperative system which uses the \ac{UAV} not only as a flying sensor but also as a device to anchor a tether on top of the cliff to assist the \ac{UGV} to climb.
To the best of our knowledge, this is the first system that uses a \ac{UAV} to attach a tether that can be used by the \ac{UGV} to extend the reachable area.

The contributions of this paper are:
\begin{enumerate}
\raggedright
\item Proposal of a novel cooperative system which:
\begin{itemize}
\item collaboratively navigate through the unknown environment,
\item uses the \ac{UAV} to attach a tether to an inaccessible place for the \ac{UGV},
\item enhances the \ac{UGV}s traversability by using the attached tether to climb a steep terrain
\end{itemize}
\item Comparison of the tether attachment methods,
\item Building a framework for autonomous collaborative navigation, cliff detection and  tether attachment.
\end{enumerate}

The remainder of this paper is organized as follows.
Section 2 describes the proposed system and mission statement.
The system overview of the \ac{UAV}, \ac{UGV} are given in Section 3 and 4 respectively.
The tether attachment methods are shown in Section 5.
Sections 6 describes our experimental set-up and results.
In Section 7, we conclude with our vision towards the future work.
\section{Proposed System}
The overview of the proposed system is described in this section.
Then, the details of an experimental mission to demonstrate the feasibility will follow.
\subsection{System overview}
The main concept of our proposed system is that a \ac{UAV} and a \ac{UGV} are connected with a tether, 
allowing the \ac{UAV} to attach the tether to a certain point, thus enhancing 
the accessibility of the \ac{UGV}.
The \ac{UAV} can be considered as a useful tool for the \ac{UGV} since it can extend the scannable and accessible area impossible to reach only with the \ac{UGV}.

\begin{figure*}[h]
	\begin{center}
		\vspace{0.5cm}
		\includegraphics[width=0.7\linewidth]{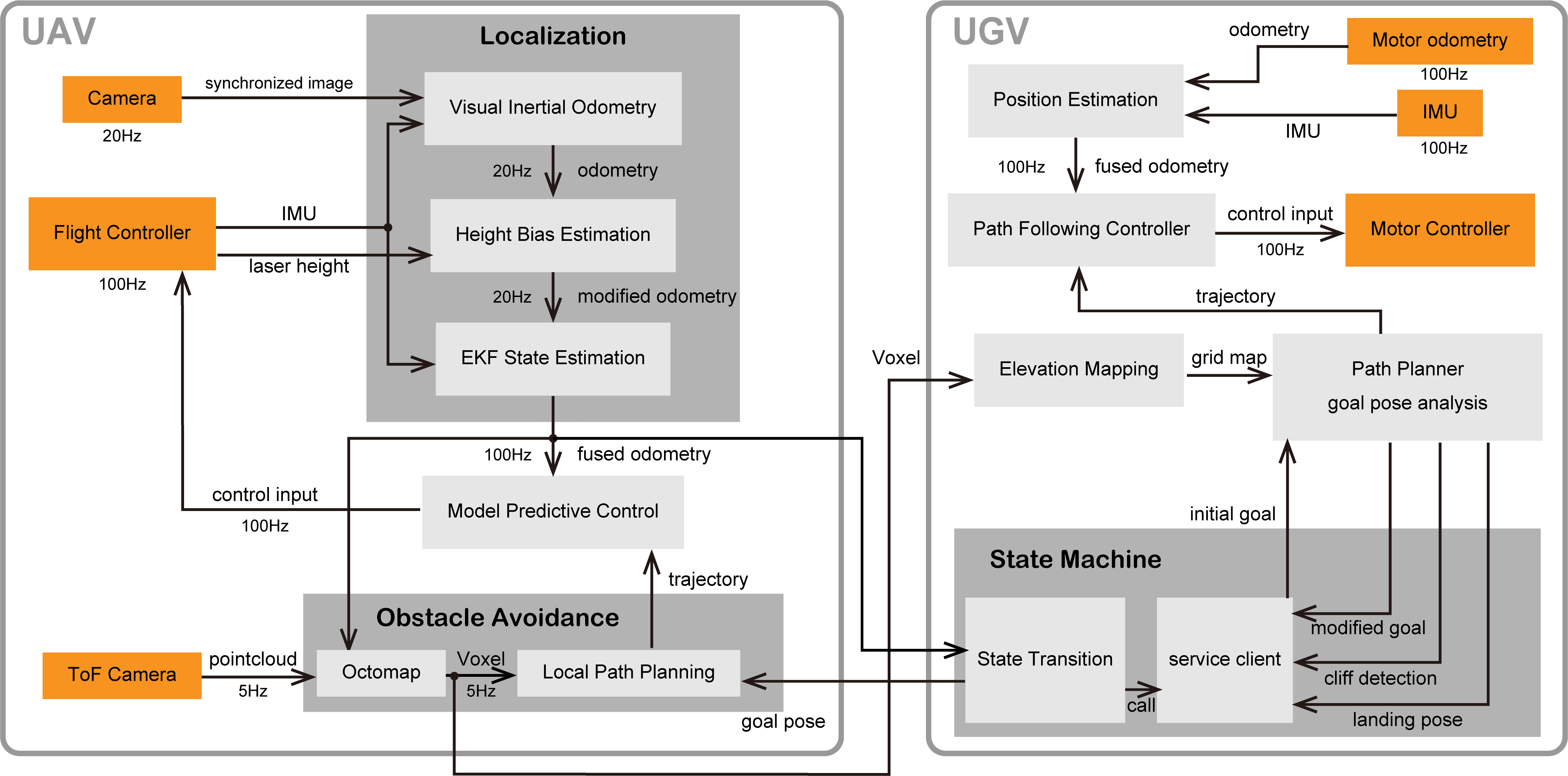}
	\end{center}
	\caption{Framework of the autonomous system based on \ac{ROS}. The computation of the \ac{UAV} is done on the Jetson TX2 and the computation of the \ac{UGV} is done on UP Core processing unit. Two robots are connected to the same network through Wifi.}
	\label{fig:framework}
\end{figure*}

To realize this system, the \ac{UAV} should be able to detect the anchoring point and automatically attach the tether to it. In addition, the \ac{UGV} should be equipped with a winch powerful enough to lift its own mass.
Furthermore, to operate autonomously, the \ac{UAV} needs a position estimation, trajectory following, obstacle avoidance and a 3D mapping capability. Also, the \ac{UGV} must plan a path based on the traversability analysis calculated from the \ac{UAV}'s sensor measurement data. 
The two robots are connected to the same network via a Wifi connection and share information such as estimated pose and generated 3D map on \ac{ROS} (Fig.\ref{fig:framework}). 
The details are described in the following sections (Section \ref{sec:uav_overview}, \ref{sec:conclusion}).

\subsection{Mission}
\label{sec:mission}
We tested the feasibility of our concept by doing an experiment of an autonomous mission.
In this mission, a \ac{UAV} and a \ac{UGV} start from a certain position and maneuver towards a goal position. Until they reach a cliff, the \ac{UAV} move towards the goal ahead of the \ac{UGV}, to create a map. The \ac{UGV} plans a path that avoids obstacles or a rough terrain using this map and follows the \ac{UAV}. Once they detect a cliff, the \ac{UAV} flies above the cliff and starts searching an anchoring point. When it finds the point, the \ac{UAV} starts flying around it to attach the tether and autonomously land to a safe area after the attachment. Then, the \ac{UGV} starts winding the tether to climb the cliff. In this experiment, the mission finishes with the successful climb of the cliff.
The tether attachment methods are explained in Section \ref{sec:tether_attachment}.

\begin{figure*}[t]
 \begin{minipage}{0.40\hsize}
  \begin{center}
   \includegraphics[width=6.0cm]{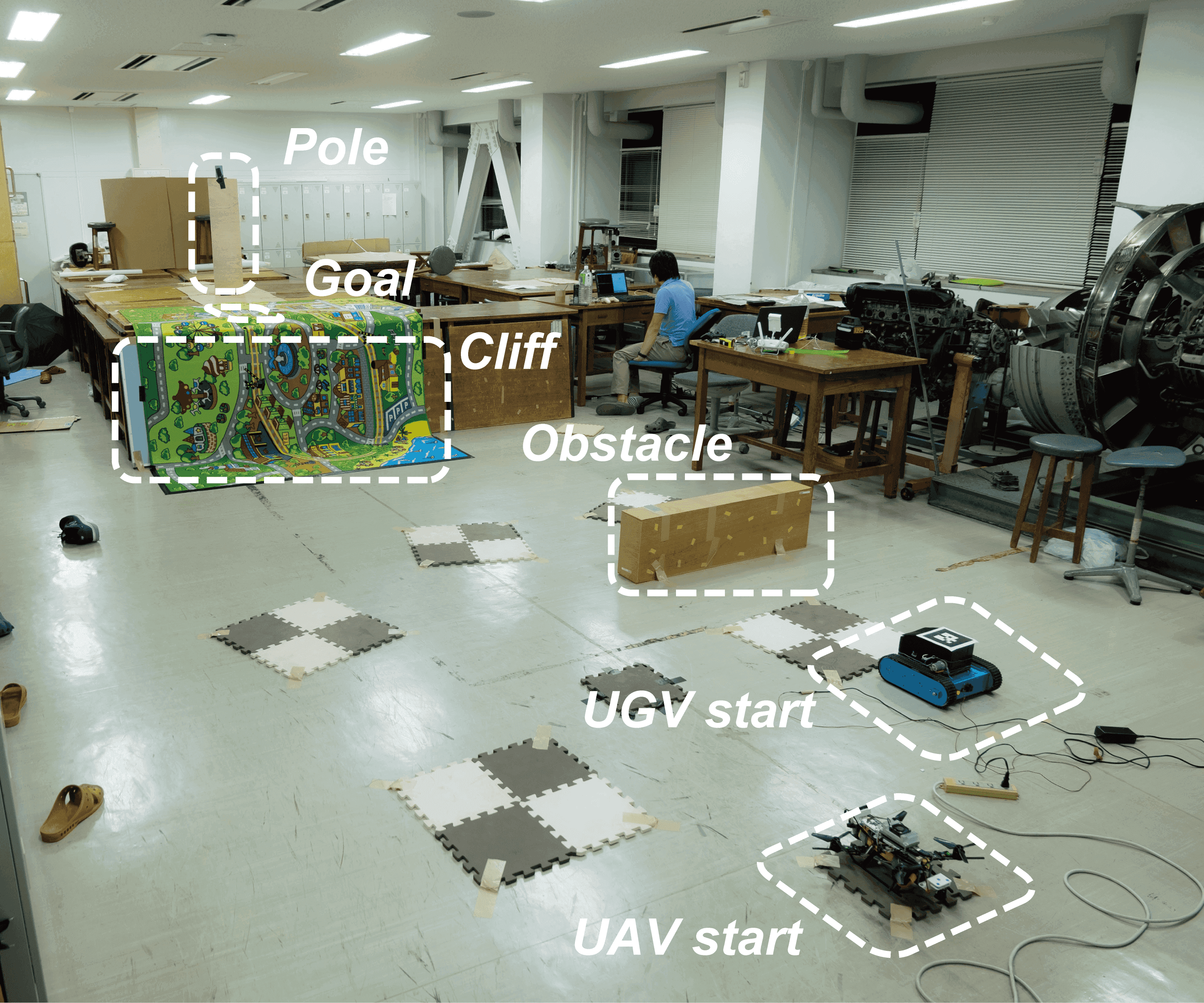}
  \end{center}
  \captionsetup{labelformat=empty,labelsep=none}
  \subcaption{The field for the whole mission}
  \label{fig:field}
 \end{minipage}
 \begin{minipage}{0.40\hsize}
  \begin{center}
   \includegraphics[width=6.0cm]{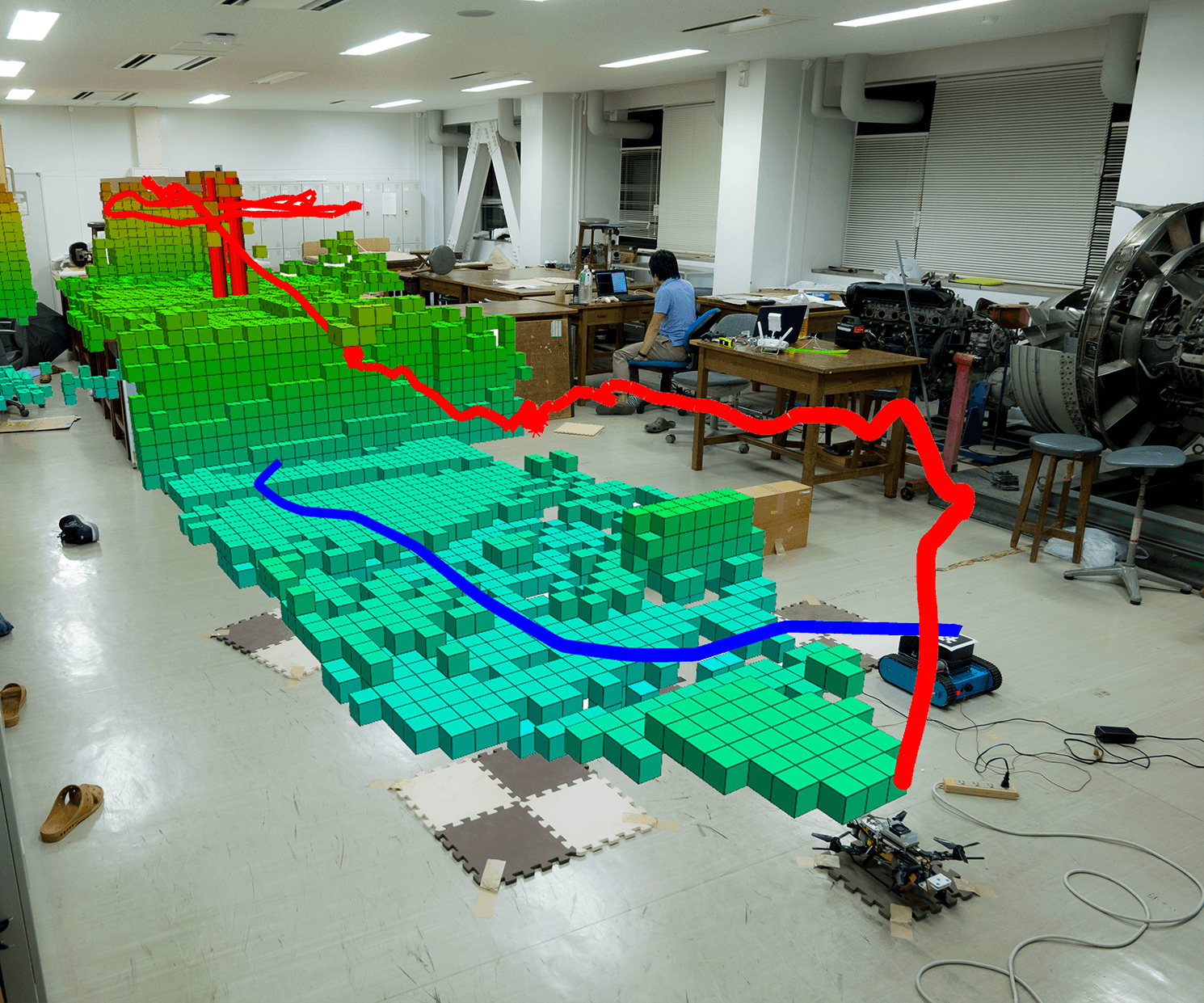}
  \end{center}
  \captionsetup{labelformat=empty,labelsep=none}
  \subcaption{3D voxel map and actual trajectory at the mission.}
 \end{minipage}
 \begin{minipage}{0.19\hsize}
  \begin{center}
   \includegraphics[width=3.5cm]{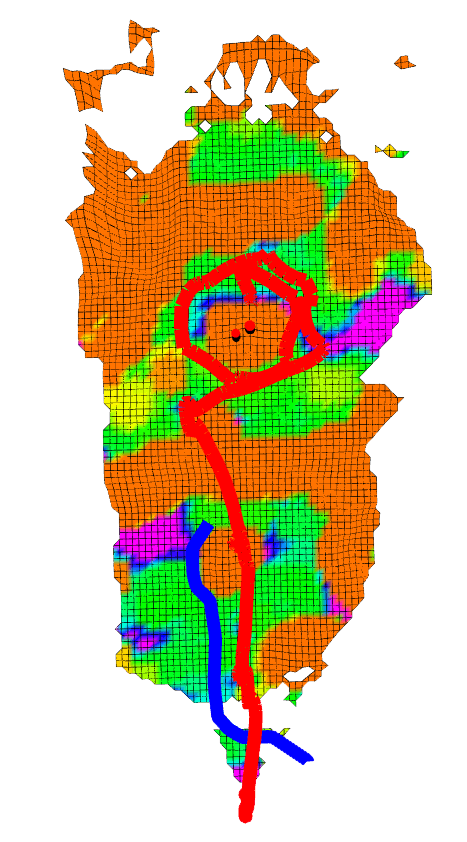}
  \end{center}
  \captionsetup{labelformat=empty,labelsep=none}
  \subcaption{Top view of the map.}
 \end{minipage}
 \caption{(a) Two robots start from the starting position. An obstacle is located in front of the \ac{UGV} and a cliff is in the middle. A pole is equipped at the top of the cliff. (b) shows the 3D voxel map after the mission execution. The red and blue line shows the actual path taken by the \ac{UAV} and \ac{UGV} respectively. The red cylinder shows the detected anchoring point. (c) shows the top view of the smoothed elevation map with a traversability color. The path of the robots are shown in the same color as (b).}
 \label{fig:whole_mission}
\end{figure*}
\section{\ac{UAV} System Overview}
\label{sec:uav_overview}
In this section, an overview of the \ac{UAV} is described.
\subsection{Hardware}
\begin{figure}[t]
 \begin{minipage}{0.49\hsize}
  \begin{center}
   \includegraphics[width=\linewidth]{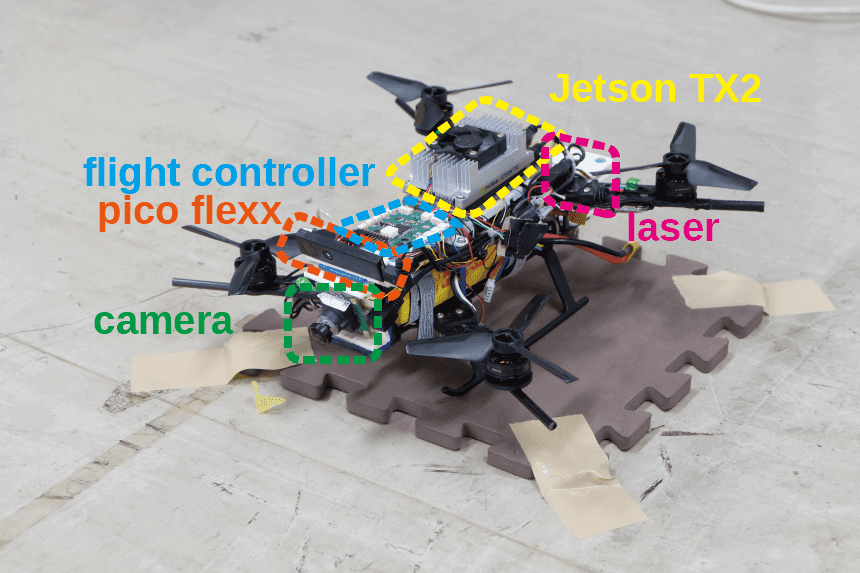}
  \end{center}
  \captionsetup{labelformat=empty,labelsep=none}
  \subcaption{Hardware of the \ac{UAV}.}
  \label{fig:uav_hardware}
 \end{minipage}
 \begin{minipage}{0.49\hsize}
  \begin{center}
   \includegraphics[width=\linewidth]{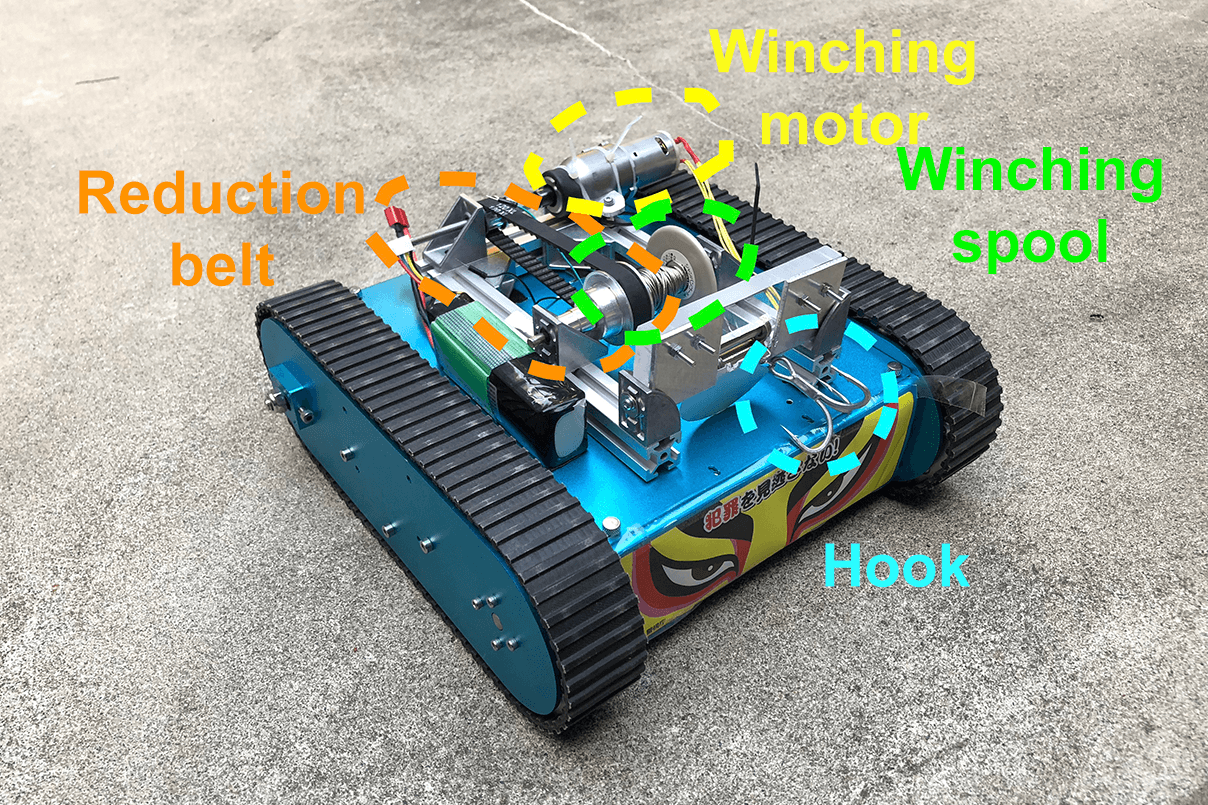}
  \end{center}
  \captionsetup{labelformat=empty,labelsep=none}
  \subcaption{Hardware of the \ac{UGV}.}
  \label{fig:ugv_hardware}
 \end{minipage}
 \caption{The hardware of the two custom-made robots.}
\end{figure}

We developed a custom-made quadrotor with NVIDIA Jetson TX2 for fully onboard localization, mapping and navigation (Fig.\ref{fig:uav_hardware}). A custom made flight controller was installed for attitude control and sensor interfaces. It has a global shutter monocular fisheye camera, a \ac{ToF} sensor and a laser sensor for height measurement. The flight controller has an \ac{IMU} for the attitude control. The frequency of each sensor measurements are image: 20hz, pointcloud: 5Hz, laser: 100Hz and \ac{IMU}: 100Hz.
The TX2 and the flight controller are connected with a serial communication link, sending telemetry at 100Hz and receiving commands at the same frequency.

\subsection{System}
The navigation of the \ac{UAV} runs on \ac{ROS}.
For the localization and position control, we followed the similar framework as \citep{sa2017build}.
The \ac{UAV} uses \ac{VIO} to estimate the relative position and orientation from the reference frame. Since this position frequency is not high enough for position control,
a \ac{MSF} framework~\citep{lynen2013robust} was used to fuse with the \ac{IMU} measurement.
In addition, as the tether attachment maneuver requires the \ac{UAV} to fly near the structure or a cliff, an obstacle avoidance trajectory planning was performed to calculate a reference trajectory for the position control framework.
For position control, a \ac{MPC} was used to calculate the roll, pitch, yaw rate and thrust command to the flight controller. The controller controls the motor speed with PID control to follow the given attitude.

\subsubsection{\ac{VIO}}
As also discussed by \citep{delmerico2018benchmark}, 
there are several open-source libraries available for \ac{VIO}~\citep{qin2017vins, bloesch2017iterated, leutenegger2015keyframe, sun2018robust}. We used ROVIO because it does not need an initialization movement and computation is relatively lightweight.
\subsubsection{Sensor Fusion}
Based on the estimated pose and velocity, we fused them with the height from the laser sensor to compensate the height drift of the \ac{VIO}. A kalman filter was used to estimate the \ac{VIO} bias, laser bias and also to filter out the outlier measurements of laser sensor based on the mahalanobis distance of the estimated variance. 
Then an \ac{EKF} was used to fuse the height filtered odometry and the \ac{IMU} measurement to acquire the high-frequency odometry estimation with the \ac{MSF} framework.
\subsubsection{Obstacle Avoidance}
For the obstacle mapping, Octomap~\cite{hornung2013octomap} was used to create a voxel grid from the pointcloud. The position of the sensor is provided by the \ac{MSF}. For the obstacle avoidance trajectory planning, we used the motion primitive approach in \cite{liu2017search}. We customized the MRSL Motion Primitive Library ROS, the open-source ROS wrapped package of the planner in \cite{liu2017search} to perform 5 Hz planning.

\subsubsection{Control}
The attitude control was done on the flight controller with a PID control. Its input is the roll and pitch angle, angular velocity around the yaw axis and the thrust. To follow the trajectory from the planner, a linear \ac{MPC} was used.
We used the framework of \cite{kamel2017model} with the system identification of the first order attitude dynamics of the PID controlled quadrotor.
\section{UGV System Overview}
\label{sec:ugv_overview}
\subsection{Hardware}
We developed a custom made ground robot based on a commercially available caterpillar platform with UP Core processing unit for onboard localization, mapping and navigation (Fig.\ref{fig:ugv_hardware}). The platform hosts \ac{IMU} with accelerometer and gyroscope for attitude estimation and a custom made winch for climbing steep terrains.
The robot also has a PID controller for each motor and tracks motor movements with encoders.
Motor for tether winding uses sensorless relay-based switch for simple control, to enable high power output.
Processing unit sends commands to the controller at 100Hz and receives odometry and attitude data at the same rate.

\subsection{System}
The on-board processing unit of the \ac{UGV} runs \ac{ROS}, which is connected to \ac{UAV} as a client.
For the localization, robot uses dead reckoning based on odometry and \ac{IMU} data similar to \cite{fuke1996dead}.
The on-board computer receives voxel data of the surroundings from the \ac{UAV} and translates it to grid map.
Furthermore, the system performs path planning and cliff detection on processed map data.
For a position control, \ac{UGV} uses Pure Pursuit algorithm~\cite{coulter1992implementation} to follow the path.

\subsubsection{Map filtering}
Before performing operations on grid map we filter it with several passes to suppress the noise of the depth measurements from the \ac{ToF} camera. 

By following the methods in \cite{Fankhauser2016GridMapLibrary, fankhauser2016collaborative},
the map was filtered sequentially by applying inpainting, slope and roughness calculation, and the traversability estimation as shown below.
$$ traversability  = \frac{1}{2} (1 - \frac{slope_{rad}}{0.6}) + \frac{1}{2} (1 - \frac{roughness}{0.1}) $$

Finally, we apply a minimum filter with a radius of 30 cm to traversability layer, to expand non-traversable spots.


\subsubsection{Path planning}
Algorithm used for path planning is A*~\cite{hart1968formal}. 
This algorithm searches for the optimal path on 2D grid with the lowest total cost, given start and goal points.
The cost at index $i$ of the grid map is defined by the weighted sum of {\it unknown area cost}, {\it elevation cost} and {\it inverse traversability} as following.

\begin{equation}
 cost_i = 
	\begin{cases} 
	\frac{W_T}{T_i + \epsilon_T} + W_E E_i  & \text{if index $i$ is valid}\\
	W_{NaN} & \text{otherwise}
	\end{cases} 
\end{equation}

where $W_{NaN}$ is cost for an unknown area, $T_i$ corresponds to traversability, $W_T$ to the weight of traversability cost component, $\epsilon_t$ as a small value to avoid zero division, $E_i$ corresponds to elevation and $W_E$ to elevations weight.\\
Proposed algorithm doesn't take into account a tether and models of UAV and UGV movements.

\subsubsection{Cliff detection}
To detect a cliff, the traversability layer was utilized with the accessibility analysis.
If a low-cost path to the goal point could be found by the path planner, it is considered as cliff-free.
On the contrary, if the cost of the path is higher than a certain threshold, the path is considered to contain a cliff. 
We perturbate the destination for the planner within a certain range around the given goal point to acquire the lowest cost path.
If the lowest path cost still exceeds the threshold, it means there is an unavoidable cliff between the start and goal point. This perturbation technique is used to avoid a false detection if the goal point is on an obstacle.
The goal point of the lowest cost path is sent to the state machine for the next maneuver.

\subsubsection{Landing Pose Search}
Landing place is chosen based on the current UAV's position and it's surroundings.
It must have defined values of grid map, difference of elevation smaller than threshold, low slope and high traversability.
We perform check of all these conditions around the current position of UAV, and if such place is found it is provided to the state machine.
\section{Tether Attachment}

\label{sec:tether_attachment}
In our proposed system, the most unique part is the anchor attachment.
The \ac{UAV} is able to attach a tether on top of a cliff and also detect the optimal place to attach it using its own sensors.
In this section, we discuss the possible options and the adopted method for our system.
\subsection{Comparison of the tether attachment methods}
There are several possible ways to attach a tether to the environment.
\subsubsection{Using a grappling hook}
\label{sec:hook}
Grappling hook is a device with multiple hooks attached to a rope, and used to temporarily secure one end of a rope.
It has been used by ancient Japanese ninjas, and sailors in naval warfare to catch a ship rigging.
This method does not need a heavy mechanism but only a lightweight hook. However, finding a right place to hook on is difficult, and the attachment is not secured meaning that it could be unhooked accidentally. 

\subsubsection{Using tether winding technique}
\label{sec:winding}
This method is also very simple.
A tether is wound around a pole-like structure a few times and secured only with its friction.
The \ac{UAV} does not need a special device to attach it but, it needs to detect a pole-like structure and fly around in an appropriate path. \cite{augugliaro2013building} used this approach to make a rope bridge but, this was done with carefully pre-calculated trajectories and an accurate positioning with a motion capture system. 

\subsubsection{Using a grasping device}
Another possible way is to use a grasping device.
The \ac{UAV} itself can become a kind of anchor by grasping the anchoring point. It is easy to unlock the gripper when the \ac{UGV} finished climbing.
However, it cannot grasp a large structure, or may become too heavy to meet the torque requirement for large \ac{UGV}s. 

\subsubsection{Inserting a peg into the ground}
Another way is to use a peg as an anchor to the ground.
This method can be used at most places where there is a soil. It does not need an anchor to attach. 
The downside of this method is that the mechanism for insertion would become complex.

\subsubsection{Using a magnet}
In \cite{zhang2017spidermav}, a \ac{UAV} attaches tethers by launching a magnetic anchor device to the target.
However, the target is limited to the magnetic surface.

\subsubsection{Hybrid of the grappling hook and winding technique}
This method can utilize the benefits of both methods.
The \ac{UAV} winds the grappling hook around the target structure. 
Compared to methods \ref{sec:hook} and \ref{sec:winding}, the chance of successful anchoring will be increased.
The overviewing comparison can be seen on the Tab. \ref{tab:comparison}.

Comparing these methods, we decided to use the hybrid method of the hook and winding technique because of its simple mechanism and the high chance of successful attachment, although, a solution to unhook the tether should be considered in future research.
To utilize this method, the \ac{UAV} must detect a suitable target for the attachment.

\begin{table}[t]
\vspace{0.5cm}
  \caption{Comparison of different anchoring methods.}
  \begin{tabular}{rcccccc}
    \hline & 1 & 2 & 3 & 4 & 5 & 6\\
    \hline
    \begin{minipage}{2mm}
      \centering
    \end{minipage} &
    \begin{minipage}{7mm}
      \centering
      \includegraphics[width=0.7cm]{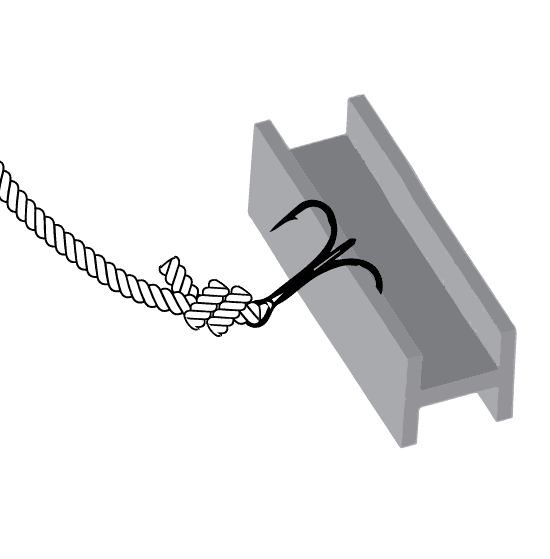}
    \end{minipage} &
    \begin{minipage}{7mm}
      \centering
      \includegraphics[width=0.7cm]{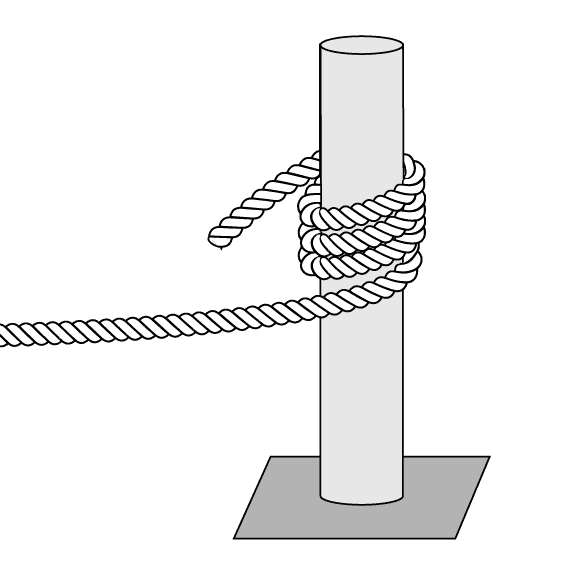}
    \end{minipage} &
    \begin{minipage}{7mm}
      \centering
      \includegraphics[width=0.7cm]{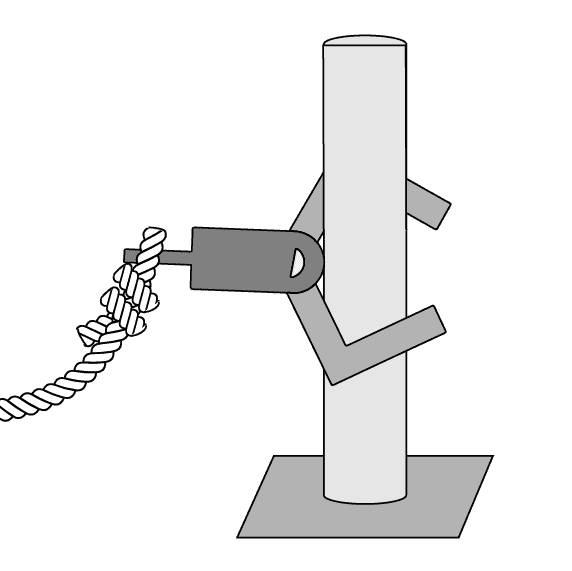}
    \end{minipage} &
    \begin{minipage}{7mm}
      \centering
      \includegraphics[width=0.7cm]{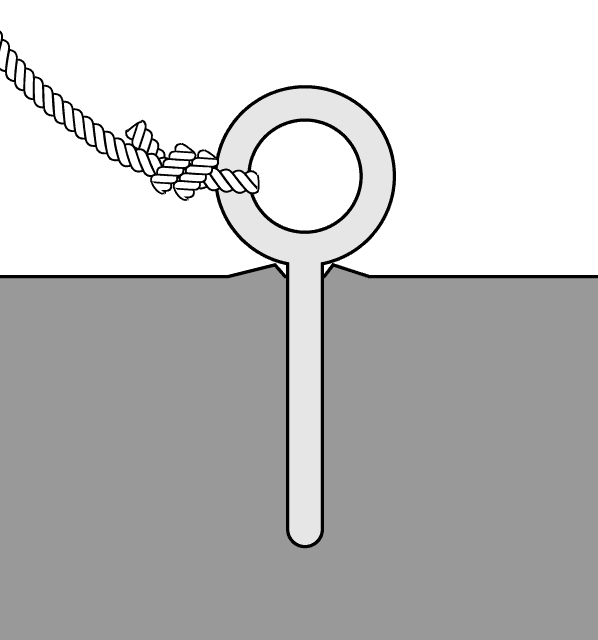}
    \end{minipage} &
    \begin{minipage}{7mm}
      \centering
      \includegraphics[width=0.7cm]{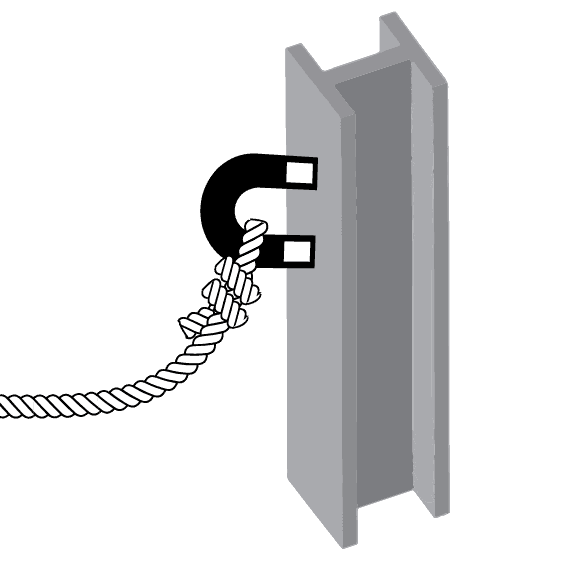}
    \end{minipage} &
    \begin{minipage}{7mm}
      \centering
      \includegraphics[width=0.7cm]{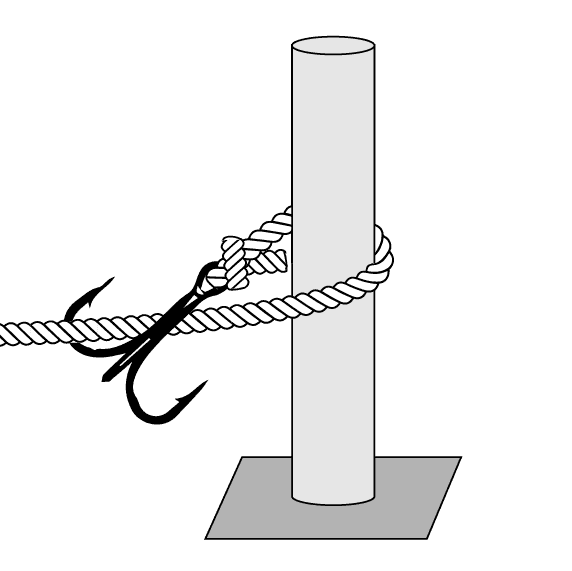}
    \end{minipage} \\ \hline
    
    device\\simplicity & *** & *** & ** & * & ** & *** \\ \hline
    control\\simplicity & ** & * & * & *** & *** & ** \\ \hline
    lightness & *** & *** & ** & * & ** & *** \\ \hline
    unhooking\\capability & * & * & *** & * & ** & * \\ \hline
    strength & ** & ** & * & ** & * & *** \\ \hline
  \end{tabular}
  \label{tab:comparison} 
\end{table}

\subsection{Anchor point detection}
To attach a tether, an acceptable anchor point should be at least some height from a surface, not in a cluttered area and peaked from the surface by a certain degree. The anchor point can be a tree trunk, stump, outcropping rock or a pole and so on.

To detect the anchor point, a grid map generated on the \ac{UGV} was used.
The area around the current \ac{UAV} position was scanned and a filter is applied to the elevation layer to calculate the {\it peakness}. 
a two dimensional gaussian distribution was fitted to a certain range around the cell.
The elevation variance around the cell is calculated as below.
\begin{eqnarray}
\sigma_x^2 &=& \frac{1}{N}\sum_{\bm{p} \in \mathcal{N}} h(\bm{p})(x - x_c)^2 \\
\sigma_y^2 &=& \frac{1}{N}\sum_{\bm{p} \in \mathcal{N}} h(\bm{p})(y - y_c)^2 \\
\sigma_{xy}^2 &=& \frac{1}{N}\sum_{\bm{p} \in \mathcal{N}} h(\bm{p})(x - x_c)(y - y_c) \\
\Sigma &=& \left(
    \begin{array}{cc}
      \sigma_x^2 & \sigma_{xy}^2\\
      \sigma_{xy}^2 & \sigma_y^2
    \end{array}
  \right)
\end{eqnarray}
where, $x_c$ and $y_c$ denotes the cell position of the center and $\mathcal{N}$ is a set of valid neighbourhood cells in a certain radius around the cell. N is the number of cells in $\mathcal{N}$ and $h(\bm{p})$ represents the relative elevation at $\bm{p}$ from the lowest cell in $\mathcal{N}$.
Then, we took the eigenvalue of $\Sigma$ to acquire the aligned variance and used the larger value, $\sigma_l^2$ for the peak calculation. This is because we want to have a peaked structure and not the edge. If we take the smaller value or the mean value for the peak calculation, the edge is also detected to be the anchor point.
Finally, the peakness is calculated as the inverse of the larger variance.
\begin{equation}
peakness = \frac{1}{\sigma_l^2}
\end{equation}
If the peakness is bigger than a certain threshold, it is detected as the anchor point.
This filter is only applied to the cells if they are the highest in a radius and the other invalid cells are ignored.

\section{Experimental Result}

We conducted several prior experiments to test each componets of the whole system as below. 
\begin{itemize}
\item \ac{UAV} navigation
\item \ac{UGV} wall and stairs climbing
\item Tether attachment validation
\end{itemize}
At last, a whole mission was conducted at the indoor field as shown in Fig. \ref{fig:field}.

\subsection{\ac{UAV} navigation}
\begin{figure}[!htbp]
 \begin{minipage}{0.32\hsize}
 \begin{center}
   \includegraphics[height=1.8cm]{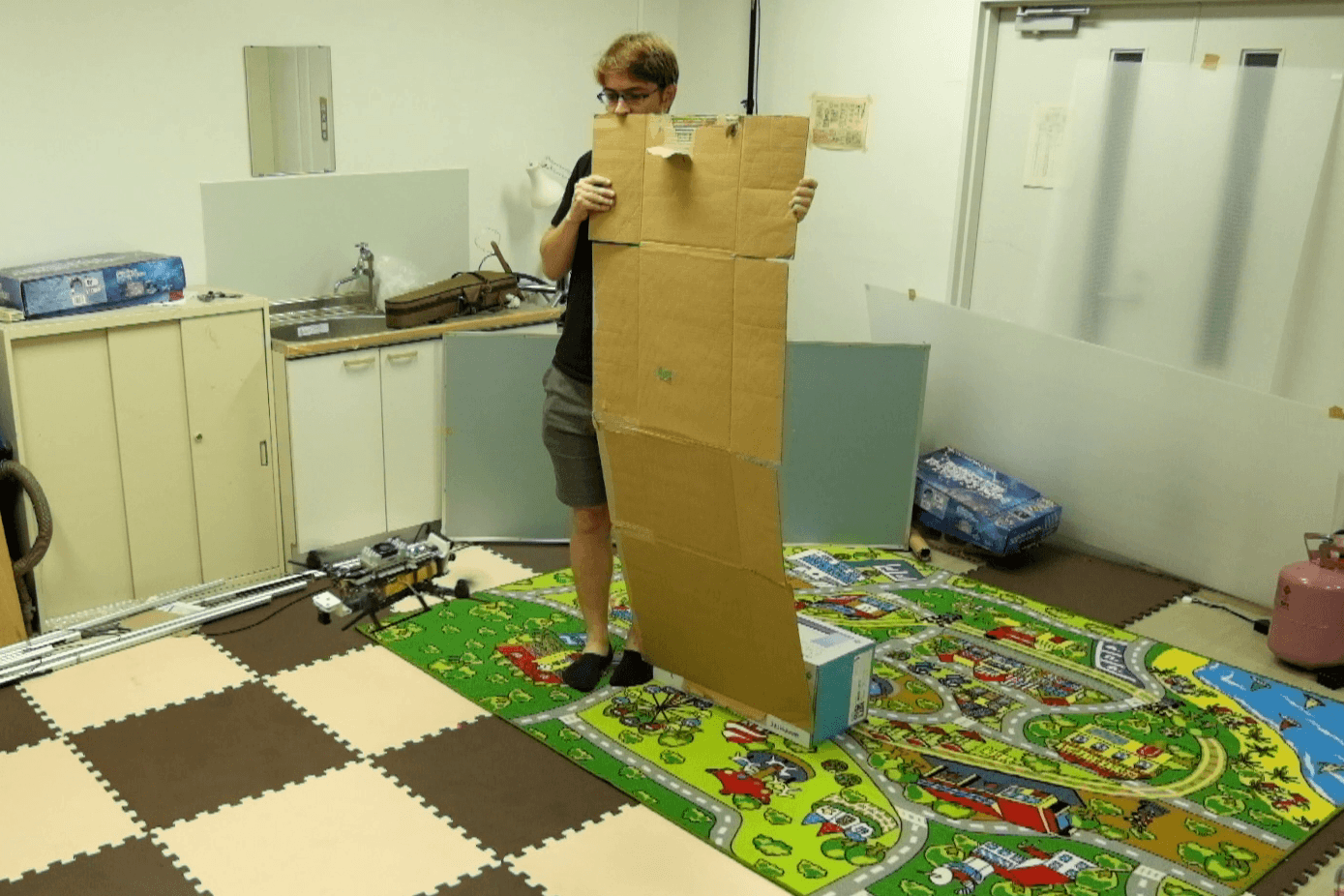}
  \end{center}
  \captionsetup{labelformat=empty,labelsep=none}
  \subcaption{}
  \label{fig:avoidance_setting}
\end{minipage}
\begin{minipage}{0.32\hsize}
\begin{center}
   \includegraphics[height=1.8cm]{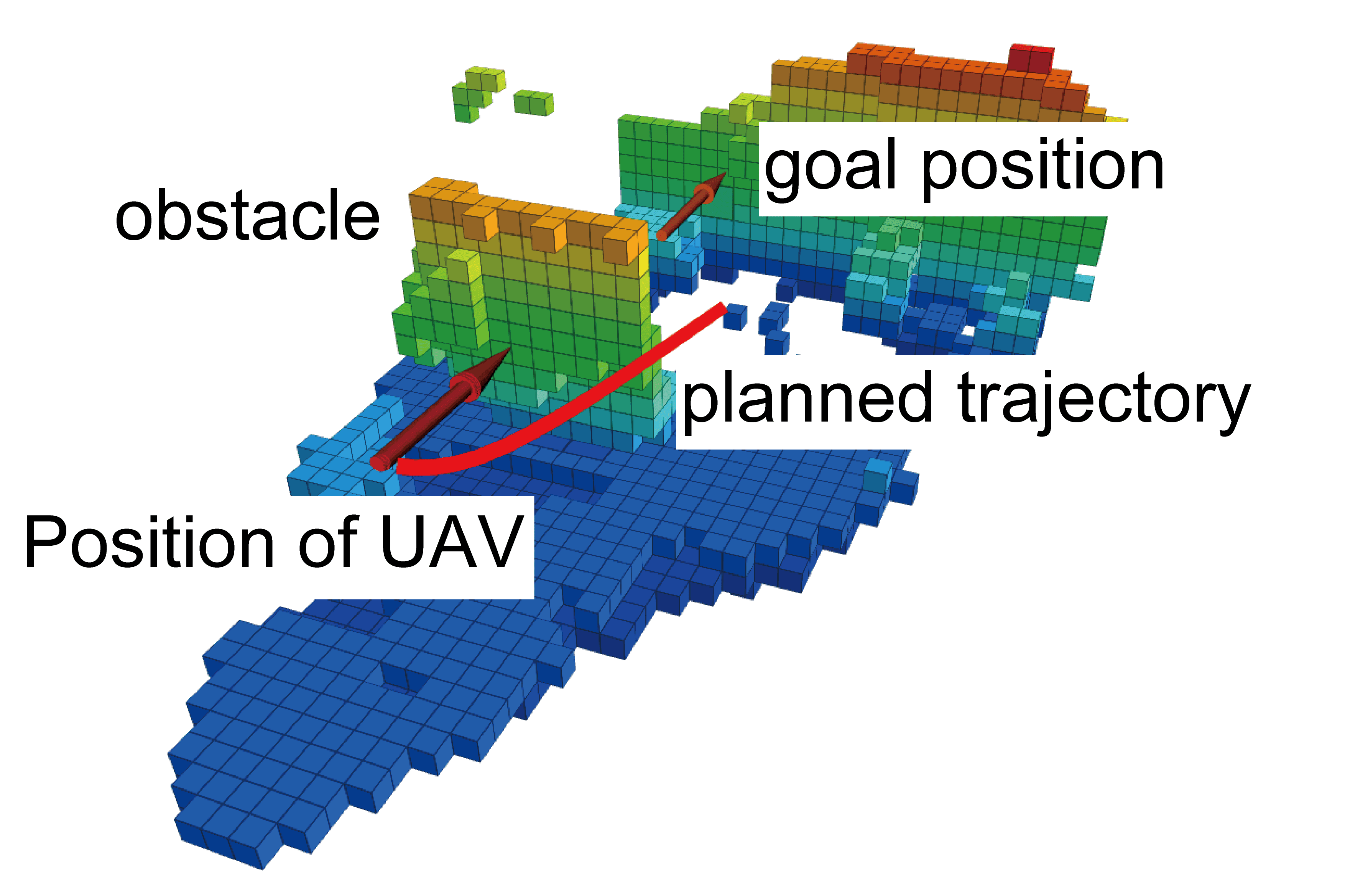}
  \end{center}
  \captionsetup{labelformat=empty,labelsep=none}
  \subcaption{}
  \label{fig:avoidance_test}
  \end{minipage}
  \begin{minipage}{0.32\hsize}
\begin{center}
   \includegraphics[height=1.8cm]{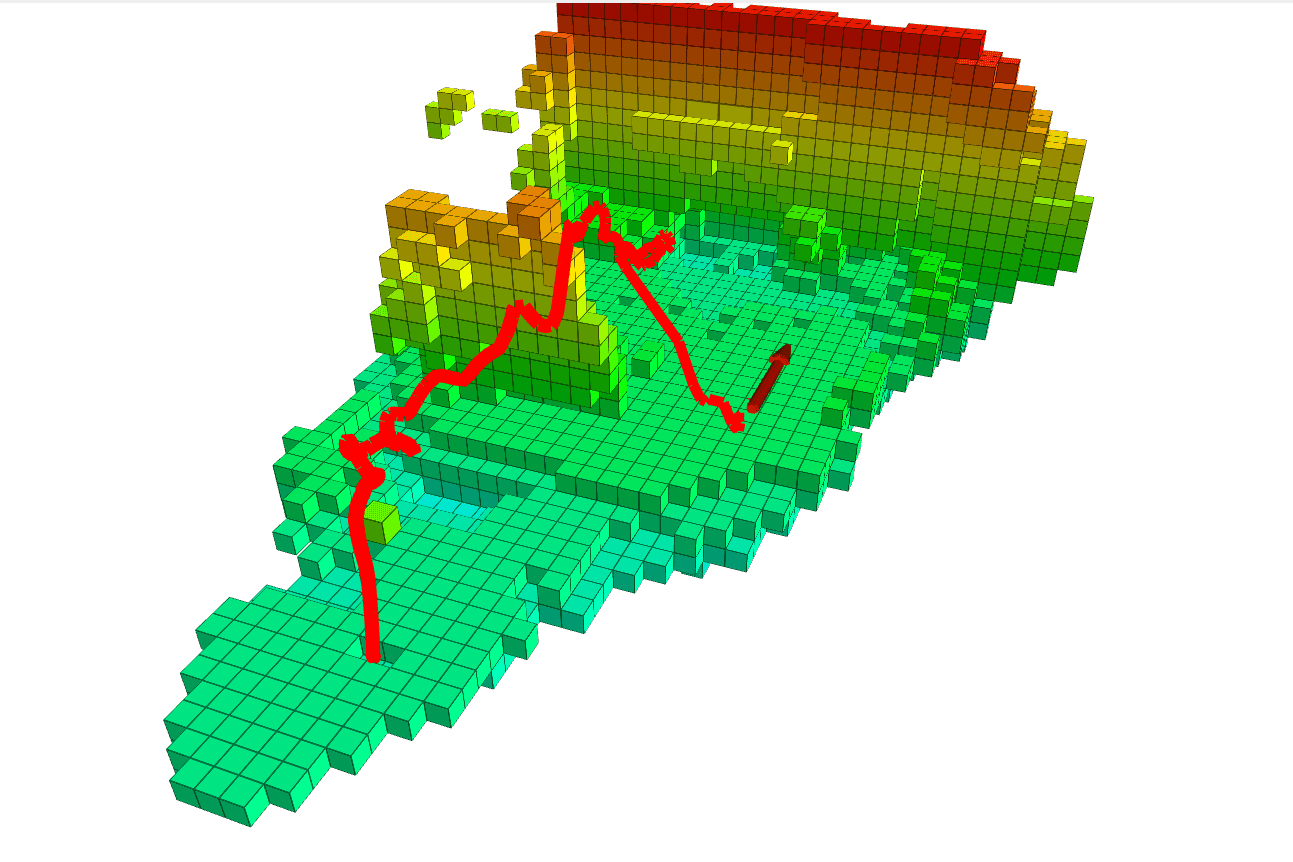}
  \end{center}
  \captionsetup{labelformat=empty,labelsep=none}
  \subcaption{}
  \label{fig:avoidance_test}
  \end{minipage}
  \caption{Obstacle avoidance experiment. (a)Shows the experimental setting. (b) visualize of the planned trajectory (c) actual trajectory of the \ac{UAV}.}
\end{figure}

In this experiment, we tested \ac{VIO} and sensor fusion, voxel mapping and local path planning.
The \ac{UAV} takes off automatically to one meter above and then, moves towards two meters in front. On the way, an obstacle held by a human appears and the \ac{UAV} must avoid this obstacle to reach the goal. At last, a grid map is used to detect a safe area for landing and the \ac{UAV} lands on the area. The final motor stop was done manually for the safety reason. The experiment setting is shown at Fig.\ref{fig:avoidance_setting}.

The result of the localization and position control was good enough to conduct a fully autonomous mission.
The \ac{UAV} successfully took off and avoided the obstacle as shown in Fig. \ref{fig:avoidance_test}.
At last, the \ac{UAV} successfully landed to the position calculated on the \ac{UGV}.

\subsection{\ac{UGV} Climbing}
\begin{figure}[!ht]
\vspace{0.2cm}
 \begin{minipage}{0.49\hsize}
  \begin{center}
   \includegraphics[width=\linewidth]{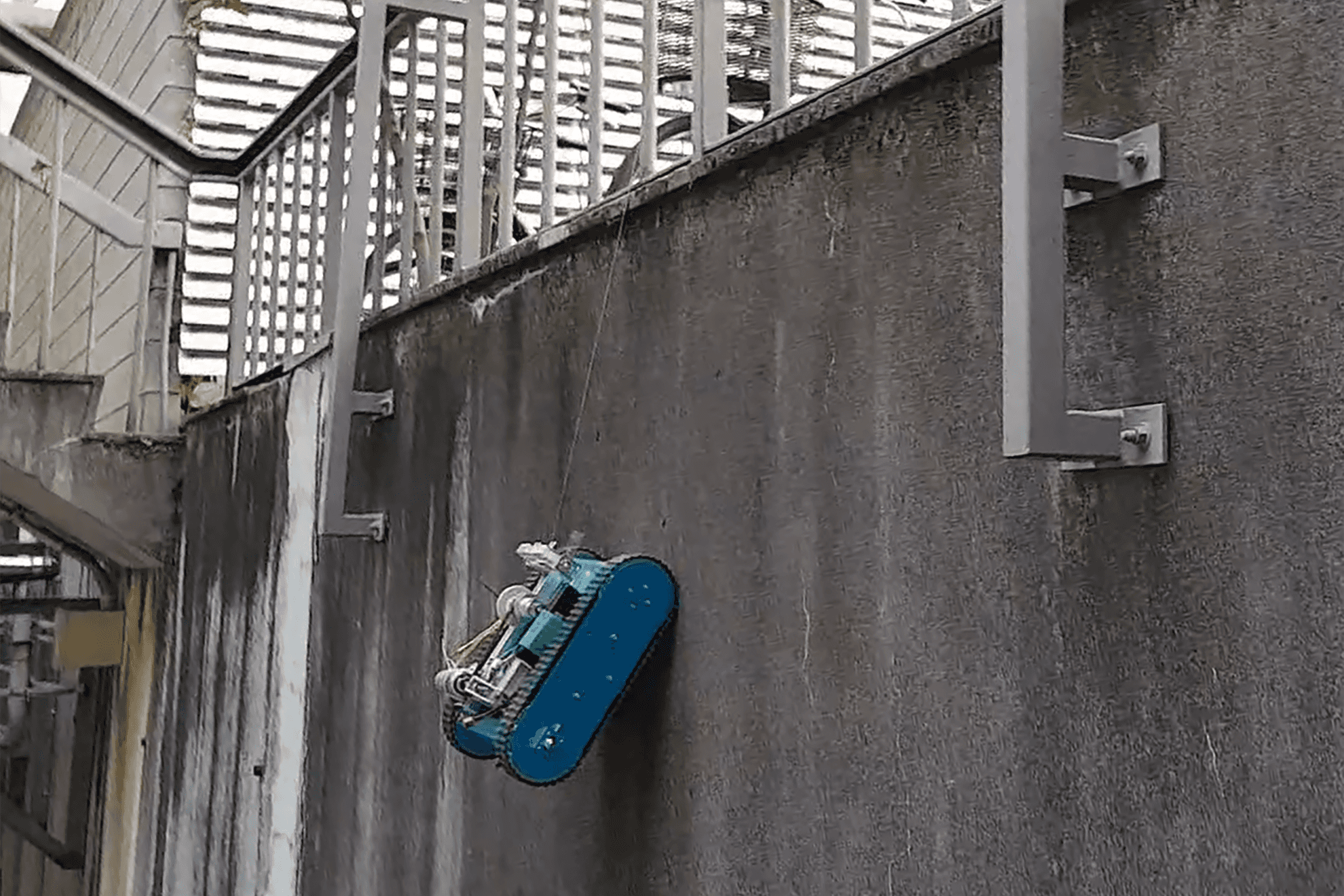}
  \end{center}
  \captionsetup{labelformat=empty,labelsep=none}
  \subcaption{Wall climbing.}
 \end{minipage}
 \begin{minipage}{0.49\hsize}
  \begin{center}
   \includegraphics[width=\linewidth]{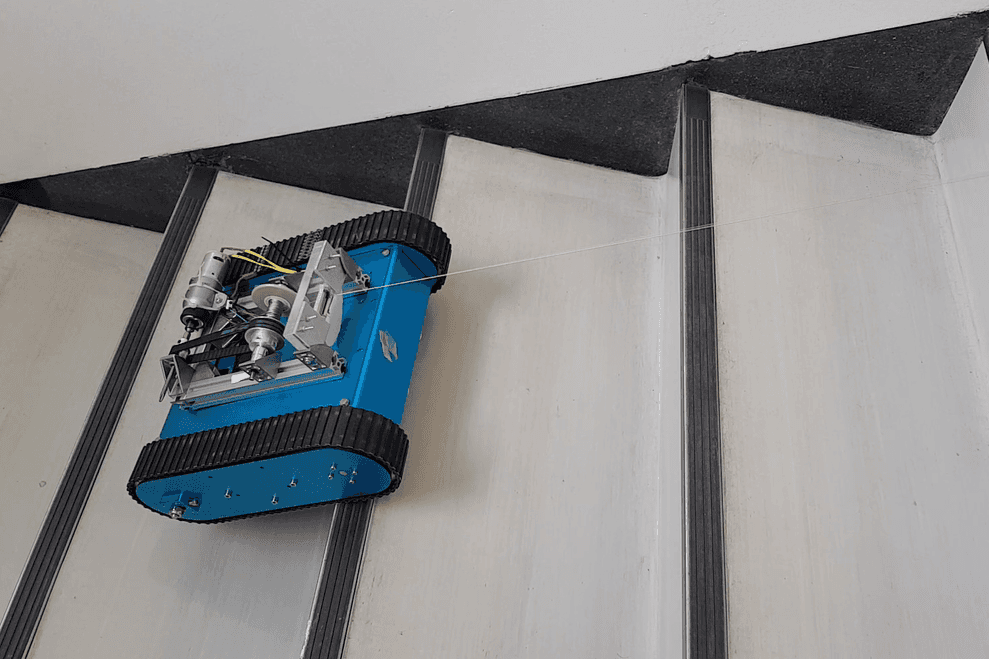}
  \end{center}
  \captionsetup{labelformat=empty,labelsep=none}
  \subcaption{Stair climbing.}
 \end{minipage} 
 \caption{Experiment of the \ac{UGV}'s climbing. The \ac{UGV} successfully climbed a vertical wall and stairs.}
 \label{fig:climbing}
\end{figure}

In this experiment, we tested the capability of the \ac{UGV}'s winch for climbing.
We attached a tether on the top of a wall and stairs by hand and winded the tether by its winch.
As shown in Fig. \ref{fig:climbing}, the \ac{UGV} successfully climbed both vertical wall and stairs with manual control.

\subsection{Tether attachment validation}
\begin{figure}[!ht]
 \begin{minipage}{0.49\hsize}
  \begin{center}
   \includegraphics[width=0.6\linewidth]{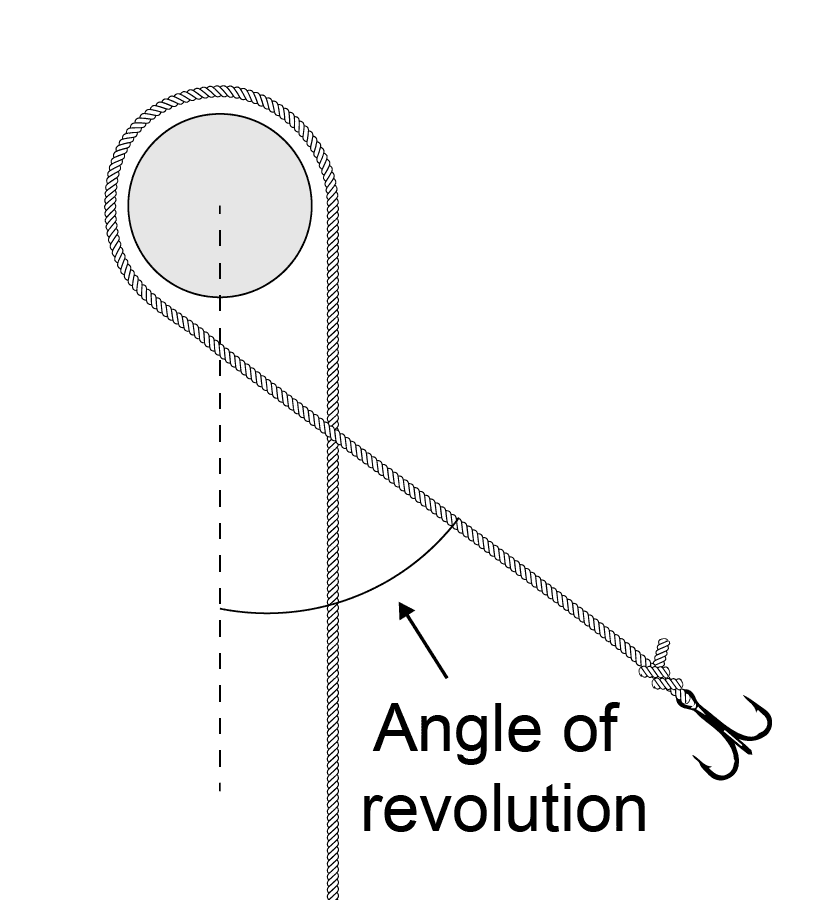}
  \end{center}
  \captionsetup{labelformat=empty,labelsep=none}
  \subcaption{Revolution angle is defined after one full revolution.}
 \label{fig:experiment_tether_setup}
 \end{minipage}
 \begin{minipage}{0.49\hsize}
  \begin{center}
   \includegraphics[width=\linewidth]{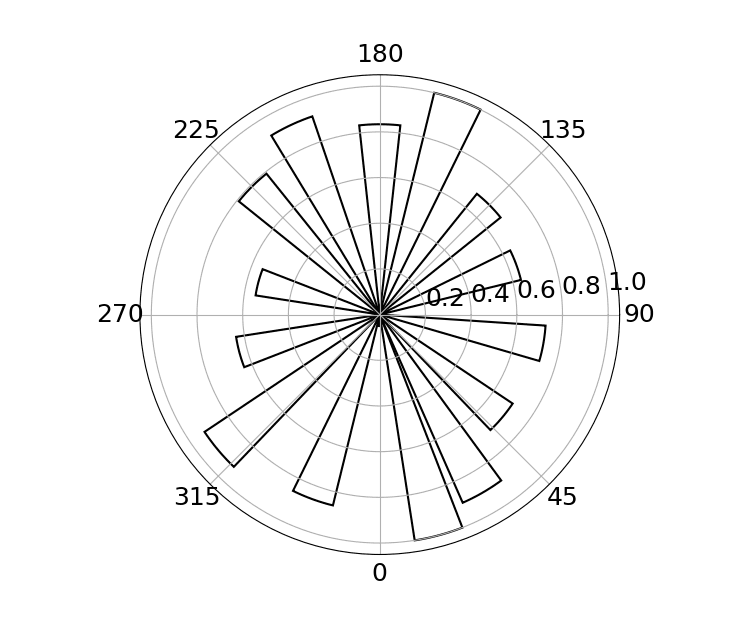}
  \end{center}
  \captionsetup{labelformat=empty,labelsep=none}
  \subcaption{Experimental results for each revolution angle.}
 \label{fig:experiment_tether_results}
 \end{minipage} 
 \caption{Tether attachment validation setup and results.}
\end{figure}
After detecting an anchor structure, the \ac{UAV} flies around it to attach a tether. 
To find an optimal angle of revolution around the pole we performed an experiment to check the probability of successful attachment at each angle after one full revolution as shown in Fig. \ref{fig:experiment_tether_setup}.
To measure the success rate, we placed a hook by hand at every 20 degrees and pulled it with the winch on a smooth flooring.
For this particular experiment, only times when the hook caught the tether was considered as a success, and therefore we counted as failures where hook got caught on the surroundings.
As shown in Fig. \ref{fig:experiment_tether_results} the most successful angle was found to be close to 0\degree or 180\degree. 
Finally, we selected 180\degree for our mission to increase the probability to be hooked to some structure and also to avoid the collision with \ac{UGV} after the climbing.

\subsection{Whole mission}
\begin{figure}[htbp]
\vspace{0.2cm}
 \begin{minipage}{0.49\hsize}
  \begin{center}
   \includegraphics[width=\linewidth]{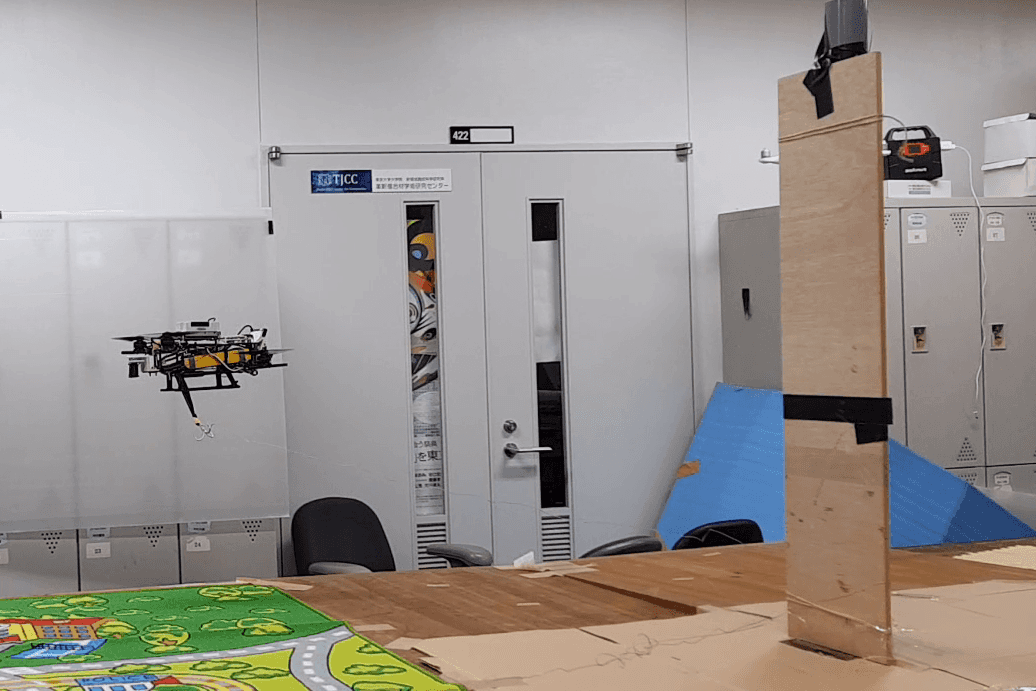}
  \end{center}
  \captionsetup{labelformat=empty,labelsep=none}
  \subcaption{The \ac{UAV} attaching a tether to a pole-like structure by flying around it.}
 \end{minipage}
 \begin{minipage}{0.49\hsize}
  \begin{center}
   \includegraphics[width=\linewidth]{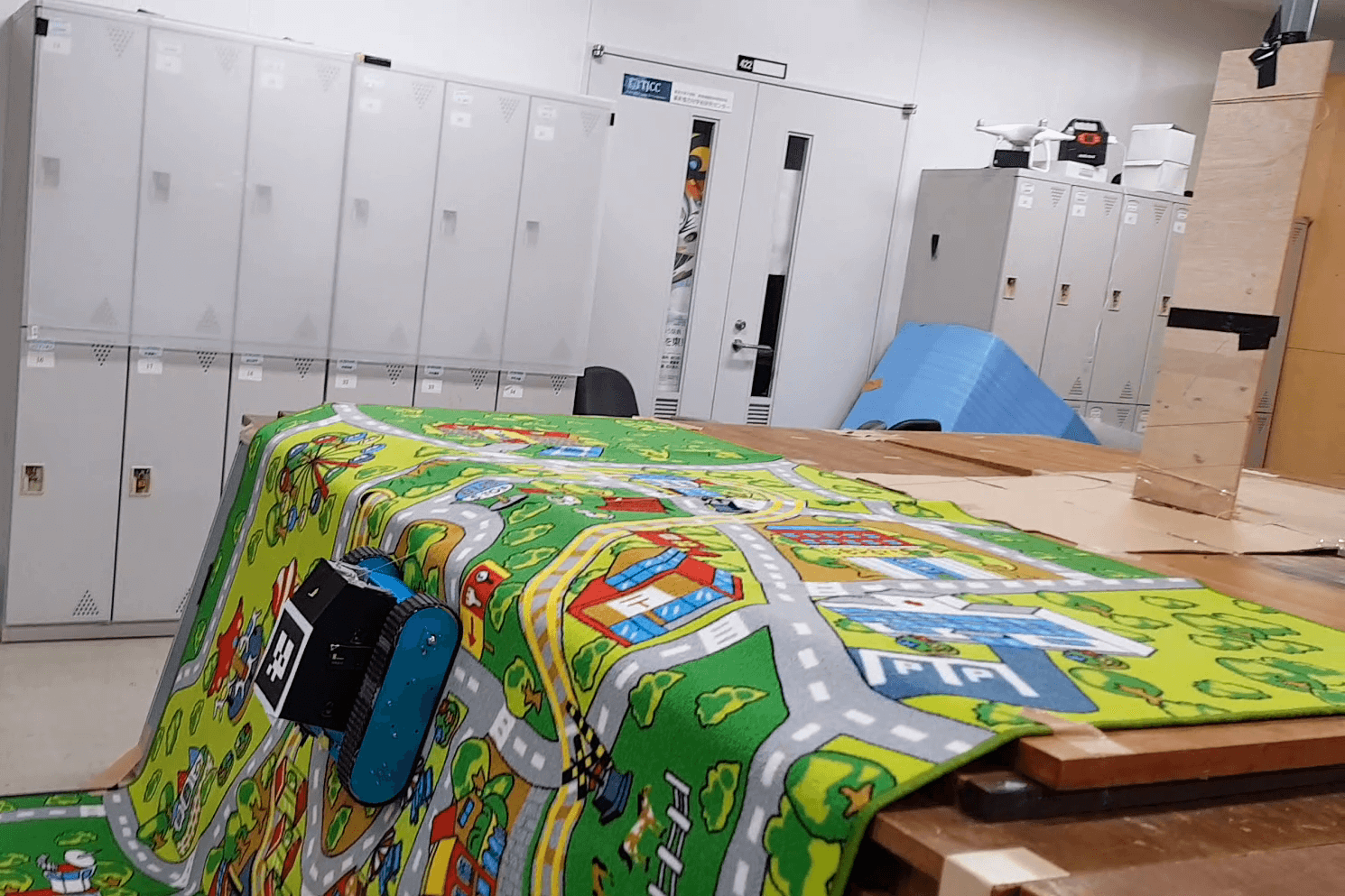}
  \end{center}
  \captionsetup{labelformat=empty,labelsep=none}
  \subcaption{The \ac{UGV} climbing a cliff by winding the tether attached by the \ac{UAV}}
 \end{minipage}
 \caption{The experimental result. The two robot successfully performed a collaborative navigation and attached a tether to a pole. Finally, the \ac{UGV} climbed the cliff by winding the attached tether.}
 \label{fig:mission_experiment}
\end{figure}
Finally, we conducted a whole autonomous mission experiment.
The field was made in an indoor room with the obstacle in front of the \ac{UGV} and a steep slope in the middle. On top of the slope, there is a pole which is intended to be an anchor point for tether attachment (Fig. \ref{fig:field}). 
As described in the mission statement section \ref{sec:mission}, the \ac{UAV} and \ac{UGV} cooperatively navigate and climb the cliff. For safety reasons, we stopped the motor of the \ac{UAV} and turned on the winch switch manually in the experiment. 

As shown in Fig. \ref{fig:whole_mission} and \ref{fig:mission_experiment}, The two robots successfully navigated through the field and reached the cliff while the \ac{UGV} avoided the obstacle. Then, the cliff was detected and the \ac{UAV} went above the cliff and began to search for a pole. After the detection of the pole, the \ac{UAV} moved to the tether attachment state. With the completion of the anchoring, the \ac{UAV} landed on a safe area and the \ac{UGV} successfully climbed the cliff.

A lesson learned from this experiment is that, a tether tension control is necessary for the tethered cooperation. Since we did not have a tension control mechanism due to the lack of sensor, the tether needed to be extended from the start and as the result, the \ac{UGV} suffered from the entangled tether many times. We observed that the navigation itself without the tether could be achieved with a high probability, however with the tether connection, the tether entangled to the \ac{UGV} as the \ac{UGV} move towards the tether on the ground, and we had to stop the experiment.

\section{CONCLUSION}
\label{sec:conclusion}
In this paper, we introduced a novel \ac{UAV} / \ac{UGV} cooperation system, which uses the \ac{UAV} not only as a flying sensor but also as a tool to attach a tether to an unreachable area for the \ac{UGV} to assist the \ac{UGV}. 
We designed the autonomous system architecture based on \ac{ROS} and open-source frameworks.
Furthermore, we compared several tether attaching methods and chose a simple hybrid method that uses a grappling hook and winding technique to increase the probability of successful anchoring. 
We conducted several experiments that include the testing of the autonomous navigation, cliff climbing and tether attachment evaluation. Lastly, we executed the whole mission autonomously. The robots successfully finished the whole mission and both robots arrived on top of the cliff thus proving the feasibility of our proposed system.

For the future work, building smarter tether tension control system is crucial.
Other directions are tether aware planning, power supply and communication via the tether, collaborative mapping and localization and tether unhooking. 


\addtolength{\textheight}{-2.0cm}  
\bibliographystyle{IEEEtranN}
\footnotesize
\bibliography{references/bibliography}

\end{document}